\documentclass[journal,twoside,web]{ieeecolor}
\usepackage{generic}
\usepackage[T1]{fontenc}
\usepackage[utf8]{inputenc}
\usepackage{cite}
\usepackage{amsmath,amssymb,amsfonts}
\usepackage{graphicx}
\usepackage{booktabs,tabularx,array}
\usepackage{stfloats}
\usepackage{textcomp}
\usepackage{url}
\usepackage{microtype}
\makeatletter
\let\NAT@parse\undefined
\makeatother
\usepackage{hyperref}
\usepackage[all]{hypcap}
\providecommand{\refname}{References}
\usepackage{etoolbox}
\patchcmd{\thebibliography}{\itemsep 0pt plus pt}{\itemsep 0pt plus 0pt}{}{\PackageError{PINR-ESI}{Bibliography spacing patch failed}{}}
\patchcmd{\thebibliography}{\section*{\color{black}}}{\section*{References}}{}{\PackageError{PINR-ESI}{Bibliography heading patch failed}{}}
\patchcmd{\subsection}{\color{subsectioncolor}}{\color{nblue}}{}{\PackageError{PINR-ESI}{Subsection color patch failed}{}}
\patchcmd{\abstract}{\color{subsectioncolor}}{\color{nblue}}{}{\PackageError{PINR-ESI}{Abstract label patch failed}{}}
\patchcmd{\abstract}{\color{subsectioncolor}}{\color{nblue}}{}{\PackageError{PINR-ESI}{Abstract label patch failed}{}}
\patchcmd{\keywords}{\color{subsectioncolor}}{\color{nblue}}{}{\PackageError{PINR-ESI}{Keyword label patch failed}{}}
\patchcmd{\keywords}{\color{subsectioncolor}}{\color{nblue}}{}{\PackageError{PINR-ESI}{Keyword label patch failed}{}}
\makeatletter
\def\fnum@figure{Fig.~\thefigure}
\def\fnum@table{TABLE~\thetable}
\long\def\@makecaption#1#2{%
  \ifx\@captype\@IEEEtablestring
    \vskip4pt
    {\normalfont\sffamily\fontsize{8}{10}\selectfont\centering\color{nblue}
     #1\par\scshape #2\par}%
    \vskip1.5pt
  \else
    \vskip\abovecaptionskip
    {\normalfont\sffamily\fontsize{8}{10}\selectfont\color{black}%
     \noindent {\color{nblue}#1.}\quad #2\par}%
  \fi}
\makeatother
\hypersetup{colorlinks=true,linkcolor=nblue,citecolor=nblue,urlcolor=nblue,pdftitle={PINR-ESI: Physics-Informed Self-Supervised Implicit Neural Representation Framework for EEG Extended-Source Imaging},pdfauthor={Zhiyong Zhou, Sheng Liao, Xiaolin Li, Youshen Xiao, Bin Dai, Surui Liu, Yakang Dai}}
\newcommand{\figref}[2][]{\hyperref[fig:#2]{Fig.~\ref*{fig:#2}#1}}
\newcommand{\figureref}[1]{\hyperref[fig:#1]{Figure~\ref*{fig:#1}}}
\newcommand{\figsref}[1]{\hyperref[fig:#1]{Figs.~\ref*{fig:#1}}}
\newcommand{\tabref}[1]{\hyperref[#1]{Table~\ref*{#1}}}
\newcommand{\eqnref}[1]{\hyperref[eq:#1]{\textup{(\ref*{eq:#1})}}}
\newcommand{\equationref}[1]{\hyperref[eq:#1]{Equation~\textup{(\ref*{eq:#1})}}}
\newcommand{\sectionref}[1]{\hyperref[#1]{Section~\ref*{#1}}}
\newcommand{\authmark}[1]{\textsuperscript{#1}}
\makeatletter
\def\ps@titlepagestyle{%
  \def\@oddhead{\vbox{%
    \hbox to\textwidth{%
      \raisebox{0pt}[8pt][0pt]{\includegraphics[height=14pt]{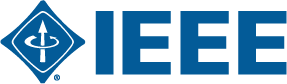}}%
      \hfil{\scriptsize\sffamily\thepage}}%
    \vskip2pt\hrule height1pt}}%
  \let\@evenhead\@oddhead
  \def\@oddfoot{}\def\@evenfoot{}}
\makeatother

\begin{document}
\title{PINR-ESI: Physics-Informed Self-Supervised Implicit Neural Representation Framework for EEG Extended-Source Imaging}
\author{Zhiyong Zhou\authmark{a,b,c,\textdagger,*}, Sheng Liao\authmark{a,b,\textdagger}, Xiaolin Li\authmark{b}, Youshen Xiao\authmark{d}, Bin Dai\authmark{b}, Surui Liu\authmark{b}, and Yakang Dai\authmark{a,b,c,*}
\thanks{\authmark{\textdagger}Zhiyong Zhou and Sheng Liao contributed equally to this work. \authmark{*}Corresponding authors: Zhiyong Zhou (e-mail: \url{zhouzy@sibet.ac.cn}) and Yakang Dai (e-mail: \url{daiyk@sibet.ac.cn}).}
\thanks{\authmark{a}The School of Biomedical Engineering (Suzhou), Division of Life Sciences and Medicine, University of Science and Technology of China, Hefei 230026, China.}
\thanks{\authmark{b}The Suzhou Institute of Biomedical Engineering and Technology, Chinese Academy of Sciences, Suzhou 215163, China.}
\thanks{\authmark{c}Key Laboratory of Biomedical Imaging Science and System, Chinese Academy of Sciences, State Key Laboratory of Biomedical Imaging Science and System.}
\thanks{\authmark{d}College of Future Technology, Peking University, Beijing, 100871, China.}}
\maketitle
\begin{abstract}
EEG extended-source imaging aims to localize active cortical regions and characterize their extent, yet supervised inverse networks typically depend on synthetic EEG--source pairs and may be sensitive to synthetic-to-real mismatch. We propose a physics-informed self-supervised implicit neural representation (INR) framework for spatial source reconstruction without paired source labels or simulation-based supervised pretraining. The lead-field matrix encodes the source-to-sensor volume-conduction model and constrains the INR through agreement between predicted and measured EEG. Cortical current density is represented as a continuous coordinate-based neural field. Forward EEG prediction samples the field at cortical vertices and applies local smoothing before lead-field mapping. A classical inverse estimate provides warm initialization and observation-specific spatial guidance through prior-guided support regularization. Simulations across varying SNRs, source extents, and source counts showed low spatial dispersion and good localization performance. Quantitative evaluation on clinical EEG recordings from epilepsy patients and qualitative analysis of scalp EEG during intracranial stimulation supported feasibility for real-world source imaging. These findings suggest the potential of physics-informed self-supervised INR for extended-source imaging.
\end{abstract}
\begin{IEEEkeywords}
Implicit Neural Representation, EEG, Inverse Problems, Neuroimaging, Source Localisation
\end{IEEEkeywords}
\section{Introduction}\label{i.-introduction}

\IEEEPARstart{E}{lectrophysiological} source imaging (ESI) estimates neural activity from noninvasive scalp EEG or MEG and provides a bridge between systems neuroscience and clinical neurophysiology \cite{ref1}, \cite{ref2}. With its high temporal resolution, ESI is valuable for epilepsy evaluation, functional mapping, and noninvasive brain monitoring. However, ESI is severely ill-posed, as multiple cortical source configurations can produce similar scalp measurements. This ambiguity is further exacerbated in extended-source imaging, where the objective is not only to localize source peaks but also to characterize the spatial distribution and extent of active cortical sources \cite{ref3}. Beyond this inherent ambiguity, supervised and conventional ESI methods face additional challenges in generalizing to real EEG and efficiently representing extended sources, respectively.

(1) Supervised inverse models are prone to a synthetic-to-real gap because real EEG--source training pairs are scarce (\figref[(a)]{1}). Most methods therefore learn from synthetic pairs generated using a forward model \cite{ref17}. These data depend on assumptions about source activity, measurement noise, and head properties \cite{ref2}, and mismatches with real recordings can limit generalization \cite{ref10}. Incorporating physics into a supervised network does not remove this dependence. For example, 3D-PIUNet refines a classical inverse estimate using a supervised network \cite{ref11}, whereas deep-unrolling methods retain forward-model operations within learned iterative updates \cite{ref9}, \cite{ref20}, \cite{ref21}. Both routes still rely on paired source labels to train their learned components. This motivates reconstruction strategies that exploit the known forward model without learning an inverse mapping from synthetic EEG--source pairs.

(2) Conventional ESI methods face a resolution-complexity trade-off imposed by predefined discrete source spaces (\figref[(b)]{1}). Cortical activity is typically represented on predefined vertices, voxels, or regions of interest over discrete meshes, so that the spatial discretization simultaneously determines the representation resolution and the number of unknown source coefficients. Coarser discretization reduces the dimensionality of the discrete source space and the number of unknowns, but its lower sampling density limits the ability to represent fine spatial variation in extended-source activity. Finer discretization improves the representation resolution and enables more detailed characterization of the source distribution; however, it increases the number of source coefficients and the size of the lead-field matrix, thereby raising computational cost and aggravating the underdetermined nature of the inverse problem.

\begin{figure}[!t]
\centering
\includegraphics[width=\columnwidth,trim={4.00bp 0.00bp 9.00bp 0.00bp},clip]{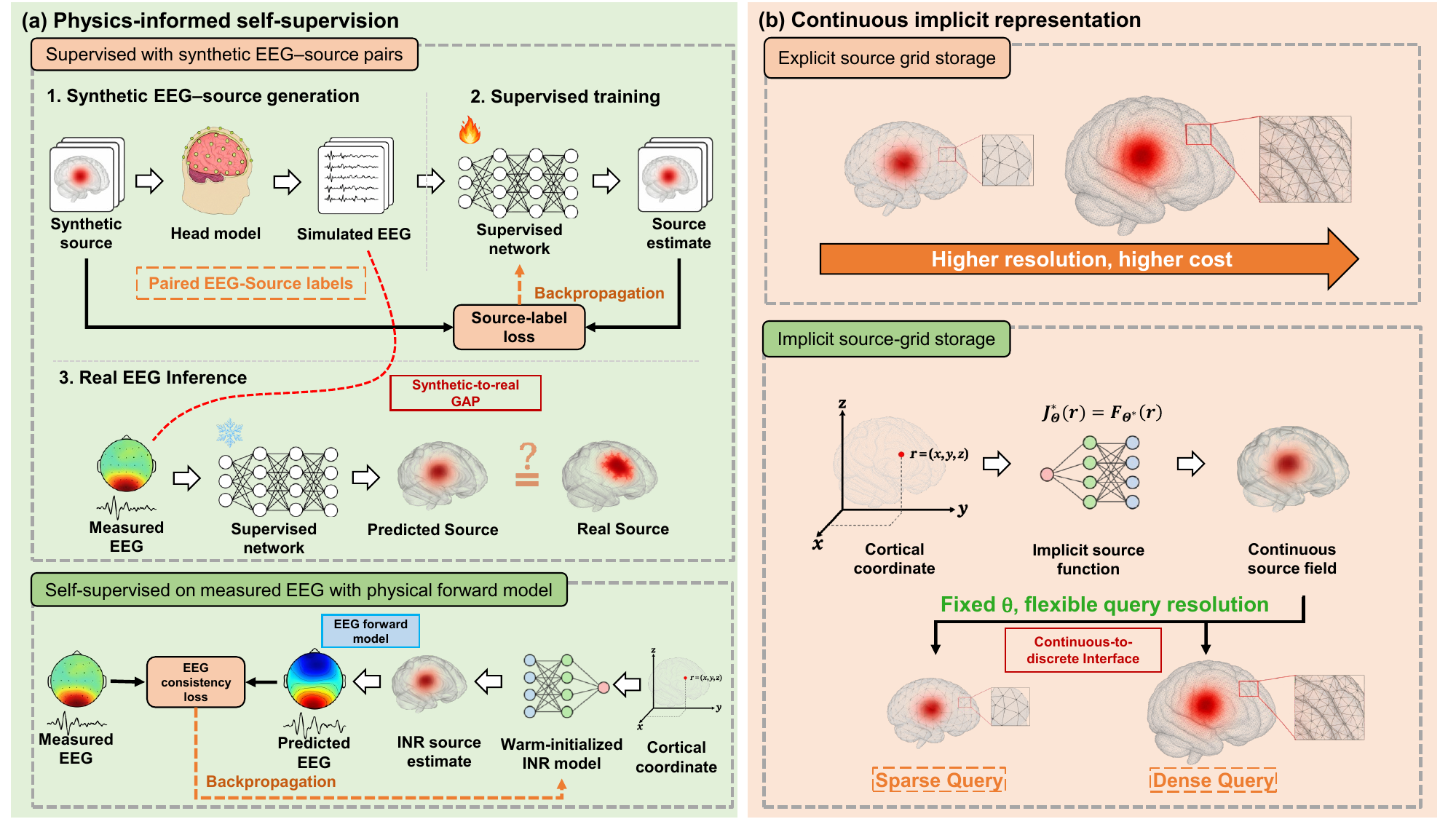}
\caption{Core concepts of PINR-ESI. (a) Unlike supervised models trained on synthetic EEG--source pairs, PINR-ESI reconstructs sources from measured EEG through forward-model consistency. (b) A continuous coordinate-based neural field represents cortical current density, enabling flexible spatial queries with shared network parameters.}
\label{fig:1}
\end{figure}

To address these challenges, we propose PINR-ESI, a physics-informed self-supervised implicit neural representation framework for EEG extended-source imaging. For the first challenge, PINR-ESI optimizes network parameters independently for each measured EEG observation rather than learning a shared inverse mapping from paired training data. The known lead-field matrix maps the reconstructed cortical sources to predicted scalp potentials \cite{ref2}, and agreement with measured EEG provides self-supervision. For the second challenge, a coordinate-based neural field parameterized by a multilayer perceptron (MLP) represents cortical current density continuously, replacing independently optimized coefficients at predefined source locations \cite{ref3}. Shared network parameters allow denser spatial queries without adding trainable source coefficients. A classical inverse estimate \cite{ref5} initializes the network and defines spatial support for prior-guided support regularization (PGSR), without serving as a ground-truth target for the final reconstruction. Forward EEG Prediction (FEP) samples the continuous field at cortical vertices, applies a fixed local smoothing operator, and maps the resulting coefficients through the lead-field matrix. Forward consistency and the spatial constraints jointly guide observation-specific optimization. The main contributions are summarized as follows.

\begin{enumerate}

\item We introduce a forward-consistency-guided self-supervised optimization strategy that removes the need for synthetic EEG--source training pairs. By reconstructing spatial source distributions directly from measured EEG, PINR-ESI avoids learning an inverse mapping from synthetic data that may not fully capture the properties of real neural activity.

\item We introduce a continuous INR source-field representation to decouple spatial sampling density from the number of trainable source parameters. The coordinate-based neural field enables flexible spatial queries with shared network parameters, avoiding a separate trainable source coefficient at each queried location.

\item We systematically evaluate PINR-ESI against representative model-based and supervised methods across varying SNRs, source extents, and source counts, demonstrating its effectiveness in identifying active cortical regions. Quantitative evaluation on clinical EEG recordings from epilepsy patients and qualitative analysis of scalp EEG recorded during intracranial stimulation further support its feasibility for real-world source imaging.

\end{enumerate}

The remainder of this paper is organized as follows. After reviewing related work in \sectionref{ii.-related-work}, we present the proposed method in \sectionref{iii.-methods} and describe the experiments in \sectionref{iv.-experimental-design}. \sectionref{v.-results} reports the results, \sectionref{vi.-discussion} discusses the findings and limitations, and \sectionref{vii.-conclusion} concludes the paper.

\section{Related Work}\label{ii.-related-work}

DeepSIF \cite{ref4} and ConvDip \cite{ref8} learn end-to-end sensor-to-source mappings, with the forward model used to generate training pairs. DST-CedNet \cite{ref22} additionally includes a forward-transformation layer and jointly minimizes source- and sensor-domain training losses. Other physics-guided designs include 3D-PIUNet, which refines an eLORETA estimate using a supervised 3D U-Net \cite{ref11}, and deep-unrolling methods such as ADMM-ESINet \cite{ref9}, XDL-ESI \cite{ref20}, and VSSI$_{2p}$-Net \cite{ref21}, which embed iterative solvers into trainable networks. Despite differences in how physics is incorporated, these approaches learn from paired synthetic data and remain dependent on the training distribution \cite{ref10}. This motivates reconstructing each observation through measured-EEG consistency without paired source-label pretraining.

Classical distributed ESI methods constrain source estimates through regularization or probabilistic priors \cite{ref2}, \cite{ref5}. MNE minimizes source energy, whereas LORETA imposes spatial smoothness through a Laplacian penalty \cite{ref5}. FAST-IRES combines iteratively reweighted edge-sparsity constraints with temporal basis functions \cite{ref6}. Block-Champagne models local source correlations using overlapping blocks within a Bayesian framework \cite{ref7}. These approaches encode different spatial or spatiotemporal assumptions, but represent source activity on predefined discrete source spaces. For a fixed spatial domain, finer source sampling increases the number of source coefficients and lead-field columns \cite{ref3}. This motivates a continuous parameterization whose trainable dimension is not tied one-to-one to query locations.

INRs model a signal or physical quantity as a continuous function evaluated at queried locations and have shown promise in solving inverse problems in medical imaging \cite{ref12}, \cite{ref16}. For ESI, this representation can describe cortical current density without assigning an independent trainable coefficient to every query location. However, the continuous neural field must still be coupled to the discrete source space of the EEG lead-field matrix. Studies of neural-field sampling and projection have examined how continuous representations can be linked to discrete measurements \cite{ref13}--\cite{ref15}. PINR-ESI couples the continuous INR source field to the discrete lead-field model through forward EEG prediction (FEP), while EEG consistency and prior-guided support regularization (PGSR) jointly guide observation-specific optimization.

\section{Methods}\label{iii.-methods}

\subsection{Overview of the Proposed Framework}\label{a.-overview-of-the-proposed-framework}

Let $\mathbf{y}\in\mathbb{R}^{N_e}$ denote a scalp EEG topography recorded by $N_e$ electrodes at a given time point, and let $\mathbf{s}\in\mathbb{R}^{N_s}$ denote the coefficients of $N_s$ cortical sources with fixed orientations. The forward model is $\mathbf{y}=\mathbf{L}\mathbf{s}+\boldsymbol{\varepsilon}$, where $\mathbf{L}\in\mathbb{R}^{N_e\times N_s}$ is the known lead-field matrix and $\boldsymbol{\varepsilon}$ denotes sensor noise.

\begin{figure*}[!tb]
\centering
\includegraphics[width=\textwidth,trim={14.00bp 4.00bp 18.00bp 4.00bp},clip]{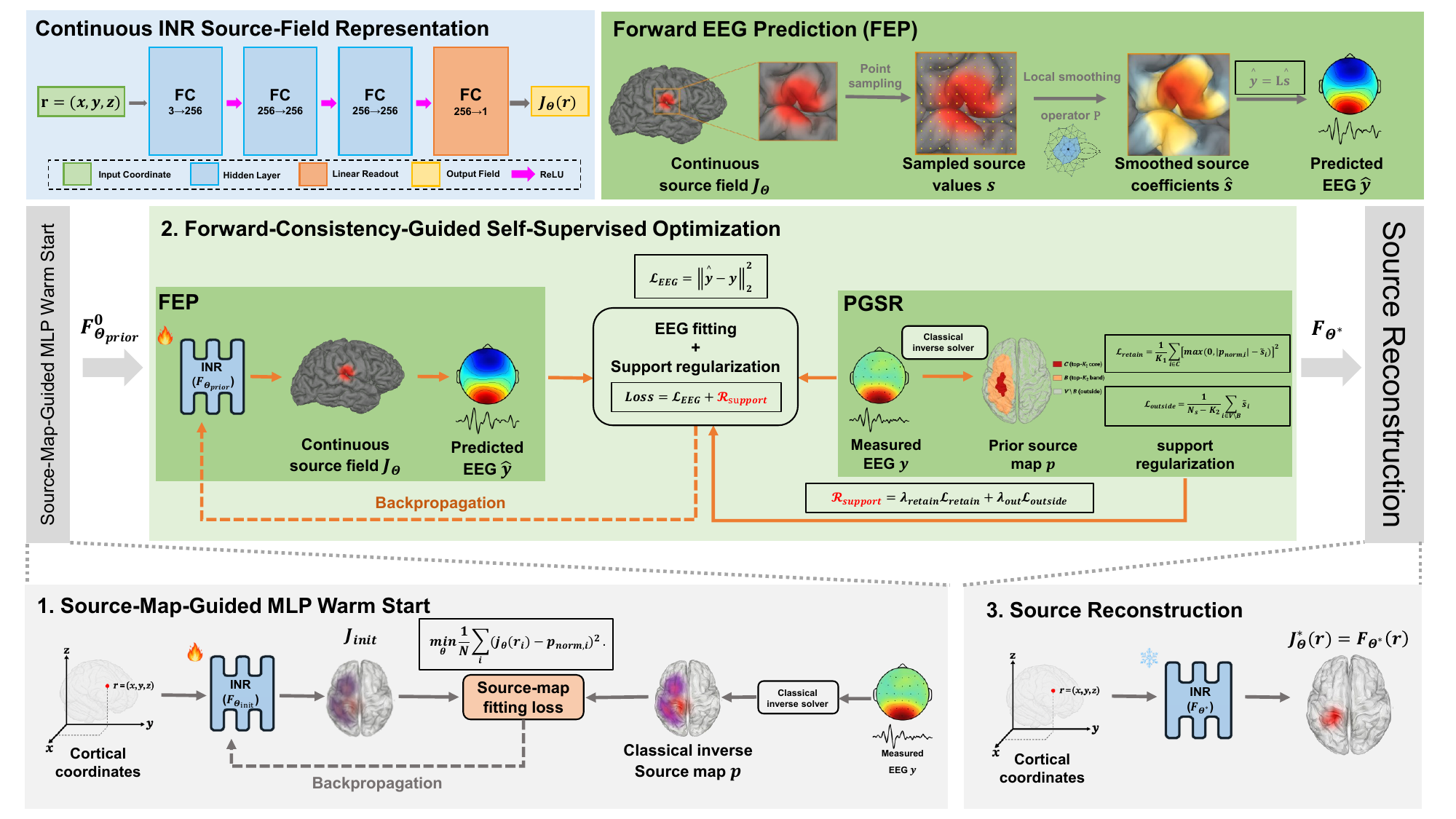}
\caption{Overall workflow of PINR-ESI. Source-Map-Guided MLP Warm Start fits a normalized classical inverse estimate to initialize the coordinate-conditioned network. During Forward-Consistency-Guided Self-Supervised Optimization, Forward EEG Prediction (FEP) samples the continuous source field at cortical vertices, applies the fixed local smoothing operator $\mathbf{P}$, and predicts EEG through the lead-field matrix $\mathbf{L}$. Prior-Guided Support Regularization (PGSR) reuses the initial estimate to encourage core activity retention and softly suppress activity outside a broader support region. EEG fitting and PGSR jointly update the network parameters. The optimized field is read out as smoothed source coefficients for reconstruction. The upper-left panel shows the MLP architecture, and the upper-right panel details FEP.}
\label{fig:2}
\end{figure*}

\figref{2} summarizes PINR-ESI for spatial reconstruction of extended cortical sources. Given $\mathbf{y}$, $\mathbf{L}$, and a cortical surface mesh $\mathcal{M}$, each time point is reconstructed independently without paired source labels or supervised pre-training. A coordinate-based neural field represents cortical current density, with one set of network parameters shared across spatial locations.

A classical inverse solver first computes a source map from the same EEG and lead field. Source-Map-Guided MLP Warm Start fits the normalized map at cortical vertices to initialize the INR parameters. The map also defines Prior-Guided Support Regularization (PGSR): a high-amplitude prior core is encouraged to retain activity, while activity outside a broader prior region is softly penalized. These constraints provide spatial guidance during refinement.

During Forward-Consistency-Guided Self-Supervised Optimization, Forward EEG Prediction (FEP) aggregates neural-field evaluations over local mesh neighborhoods using fixed geometric weights. The resulting coefficients are mapped through the lead-field matrix to predict EEG. The EEG discrepancy and PGSR jointly update the network parameters; the prior map, local smoothing operator, and lead-field matrix remain fixed. After optimization, the locally smoothed source coefficients are used for reconstruction and quantitative evaluation.

\subsection{Source-Map-Guided MLP Warm Start}\label{b.-source-map-guided-mlp-warm-start}

Let $V=\{1,\ldots,N_s\}$ index the cortical mesh vertices, with coordinates $\mathbf{r}_i\in\mathcal{M}\subset\mathbb{R}^3$. To initialize the INR parameters, we first compute a classical inverse source map. Exact low-resolution brain electromagnetic tomography estimates $\mathbf{p}\in\mathbb{R}^{N_s}$ from the same $\mathbf{y}$ and $\mathbf{L}$. The map is normalized at each vertex:

\begin{equation}
p_{\mathrm{norm},i}=\frac{p_i}{\max_{j\in V}|p_j|+\delta},\quad i\in V.
\label{eq:1}
\end{equation}

Here, $p_i$ is the estimate at vertex $i$, and $\delta$ is a positive numerical stability constant. The normalized map $\mathbf{p}_{\mathrm{norm}}$ serves as the fitting target for initialization. The network receives cortical coordinates as inputs and learns to reproduce the corresponding normalized source values.

The coordinate-conditioned multilayer perceptron (MLP) $F_\theta$ has three 256-unit hidden layers with rectified linear unit (ReLU) activations and a scalar linear output. Its mapping is $F_\theta:\mathcal{M}\to\mathbb{R}$, $\mathbf{r}\mapsto j_\theta(\mathbf{r})$, where $j_\theta(\mathbf{r})$ denotes the scalar current-density value along the prescribed source orientation. To initialize the weights and biases $\theta$, we minimize the mean squared error between network outputs and the normalized prior map at mesh vertices:

\begin{equation}
\theta_{\mathrm{prior}}=\operatorname*{arg\,min}_{\theta}\frac{1}{N_s}\sum_{i=1}^{N_s}\bigl(j_\theta(\mathbf{r}_i)-p_{\mathrm{norm},i}\bigr)^2.
\label{eq:2}
\end{equation}

This fitting step transfers the spatial pattern of the classical estimate into the network parameters. The fitted parameters $\theta_{\mathrm{prior}}$ provide the initial network weights and biases for EEG-consistency optimization. During refinement, $\theta$ is updated using measured EEG and the soft spatial constraints defined in \sectionref{c.-forward-consistency-guided-self-supervised-optimization}, allowing the source amplitudes to depart from the initialization.

\subsection{Forward-Consistency-Guided Self-Supervised Optimization}\label{c.-forward-consistency-guided-self-supervised-optimization}

Forward-Consistency-Guided Self-Supervised Optimization (\figref{2}) combines FEP and PGSR within EEG-consistency optimization. FEP generates predicted EEG from the continuous source field, while PGSR uses the classical source map to guide its spatial support. Together, they guide observation-specific refinement of the network parameters.

\subsubsection{Forward EEG Prediction (FEP)}\label{sec:fep}

The INR represents a continuous current-density field, whereas the lead-field matrix operates on a discrete cortical source space. To connect the field to the forward model, we evaluate the network at the $N_s$ source vertices in vertex order:

\begin{equation}
\mathbf{s}=\bigl[j_\theta(\mathbf{r}_1),\ldots,j_\theta(\mathbf{r}_{N_s})\bigr]^\top\in\mathbb{R}^{N_s}.
\label{eq:field-samples}
\end{equation}

The entry $s_i$ equals $j_\theta(\mathbf{r}_i)$. Although point-wise source samples can be directly mapped to EEG, this readout does not explicitly account for the local homogeneity typically assumed for extended cortical sources. We therefore introduce a fixed local smoothing operator $\mathbf{P}\in\mathbb{R}^{N_s\times N_s}$ to aggregate neighboring source values. For each vertex $i$, let $N_i^{(2)}$ contain the vertex itself and its two-ring mesh neighbors, reachable by following at most two mesh edges. The weight $P_{ij}$ is zero outside $N_i^{(2)}$ and proportional to the surface area associated with vertex $j$ within this neighborhood. The weights are normalized to sum to one for each $i$, so each row of $\mathbf{P}$ defines an area-weighted average. The operator is computed from the mesh geometry before optimization and kept fixed.

Applying $\mathbf{P}$ yields the locally smoothed source coefficients:

\begin{equation}
\hat{\mathbf{s}}=\mathbf{P}\mathbf{s}\in\mathbb{R}^{N_s}.
\label{eq:local-smoothing}
\end{equation}

Both vectors contain one scalar entry per cortical vertex, preserving the dimension of the source space. The component-wise form is

\begin{equation}
\hat{s}_i=\sum_{j\in N_i^{(2)}}P_{ij}j_\theta(\mathbf{r}_j),\quad i\in V.
\label{eq:smoothing-component}
\end{equation}

The resulting coefficients predict EEG through the lead-field matrix:

\begin{equation}
\hat{\mathbf{y}}=\mathbf{L}\hat{\mathbf{s}}=\mathbf{L}\mathbf{P}\mathbf{s}\in\mathbb{R}^{N_e}.
\label{eq:eeg-prediction}
\end{equation}

\equationref{eeg-prediction} connects coordinate-conditioned field evaluations to sensor prediction. The smoothed coefficients $\hat{\mathbf{s}}$ provide the reconstruction output; direct point-wise readout is retained for diagnostic comparisons and ablations. Denser queries increase the sampling density of the continuous field, while the physical resolution remains governed by the discrete source space associated with $\mathbf{L}$.

\subsubsection{Prior-Guided Support Regularization (PGSR)}\label{sec:pgsr}

The EEG inverse problem is nonunique, so EEG consistency alone does not uniquely determine the source distribution. Without ground-truth source labels, PGSR reuses the classical inverse estimate $\mathbf{p}$ from the same EEG observation for observation-specific spatial guidance. This estimate supports warm initialization (\mbox{\sectionref{b.-source-map-guided-mlp-warm-start}}) and defines fixed support regions for subsequent refinement.

Here, spatial support means the cortical region regarded as active, represented on the mesh by a set of vertex indices. To construct the prior regions, the indices in $V$ are ranked by $|p_{\mathrm{norm},i}|$. The core $C$ contains the top-$K_1$ indices, and the broader region $B$ contains the top-$K_2$ indices, with $C\subset B\subset V$ and $K_1<K_2<N_s$. These regions are fixed during refinement. At vertex $i$, the current field magnitude is $|s_i|=|j_\theta(\mathbf{r}_i)|$. We normalize these unsmoothed magnitudes by their maximum over all vertices, with $\delta$ ensuring numerical stability:

\begin{equation}
\widetilde{s}_i=\frac{|s_i|}{\max_{j\in V}|s_j|+\delta},\quad i\in V.
\label{eq:field-magnitude}
\end{equation}

The retention penalty discourages the normalized field amplitude from falling below the normalized prior amplitude within $C$:

\begin{equation}
\mathcal{L}_{\mathrm{retain}}=\frac{1}{K_1}\sum_{i\in C}\left[\max\bigl(0,|p_{\mathrm{norm},i}|-\widetilde{s}_i\bigr)\right]^2.
\label{eq:retain}
\end{equation}

The outside-region penalty averages the normalized field magnitude over the $N_s-K_2$ vertices outside $B$:

\begin{equation}
\mathcal{L}_{\mathrm{outside}}=\frac{1}{N_s-K_2}\sum_{i\in V\setminus B}\widetilde{s}_i.
\label{eq:outside}
\end{equation}

Rather than enforcing full-map agreement, PGSR encourages retention of prior-supported core activity with the one-sided penalty in \eqnref{retain} and softly suppresses activity outside the broader support region through \eqnref{outside}. This permits EEG-consistent refinement without forcing agreement with the initial estimate.

Combining EEG fitting from FEP with spatial regularization from PGSR yields the unified self-supervised objective. EEG fitting uses the smoothed source coefficients:

\begin{equation}
\mathcal{L}_{\mathrm{EEG}}(\theta)=\|\hat{\mathbf{y}}-\mathbf{y}\|_2^2=\|\mathbf{L}\mathbf{P}\mathbf{s}-\mathbf{y}\|_2^2.
\label{eq:eeg-fitting}
\end{equation}

The support regularization combines the penalties on the unsmoothed field:

\begin{equation}
\mathcal{R}_{\mathrm{support}}(\theta)=\lambda_{\mathrm{retain}}\mathcal{L}_{\mathrm{retain}}(\theta)+\lambda_{\mathrm{out}}\mathcal{L}_{\mathrm{outside}}(\theta),
\label{eq:support}
\end{equation}

where the nonnegative weights $\lambda_{\mathrm{retain}}$ and $\lambda_{\mathrm{out}}$ balance the two penalties. The joint optimization is

\begin{equation}
\begin{aligned}
\mathcal{L}(\theta)&=\mathcal{L}_{\mathrm{EEG}}(\theta)+\mathcal{R}_{\mathrm{support}}(\theta),\\
\theta^\star&=\operatorname*{arg\,min}_{\theta}\mathcal{L}(\theta).
\end{aligned}
\label{eq:objective}
\end{equation}

Starting from $\theta_{\mathrm{prior}}$, PINR-ESI optimizes the network parameters independently for each EEG observation. The measured EEG $\mathbf{y}$, lead-field matrix $\mathbf{L}$, local smoothing operator $\mathbf{P}$, prior map, and prior-defined support regions remain fixed. The smoothed source estimate $\hat{\mathbf{s}}$ obtained at $\theta^\star$ is used for reconstruction and quantitative evaluation.

\section{Experimental Design}\label{iv.-experimental-design}

\subsection{Experimental Setup}

\begin{figure*}[!tb]
\centering
\includegraphics[width=\textwidth,trim={4.99bp 4.98bp 0.00bp 2.98bp},clip]{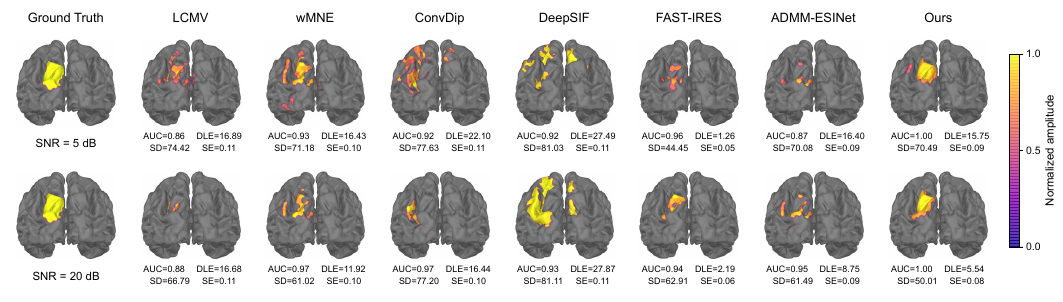}
\caption{ESI results of seven methods at SNRs of 5 dB (top) and 20 dB (bottom), with identical single-source ground truth of 10 cm$^2$. Both examples share the same single-source ground truth with a target area of 10 cm$^2$. Source maps are independently normalized to 0--1. AUC, DLE (mm), SD (mm), and SE (full-map NRMSE) are shown below each reconstruction.}
\label{fig:3}
\end{figure*}

\begin{figure}[!tb]
\centering
\includegraphics[width=\columnwidth,trim={11.00bp 0.00bp 0.00bp 0.00bp},clip]{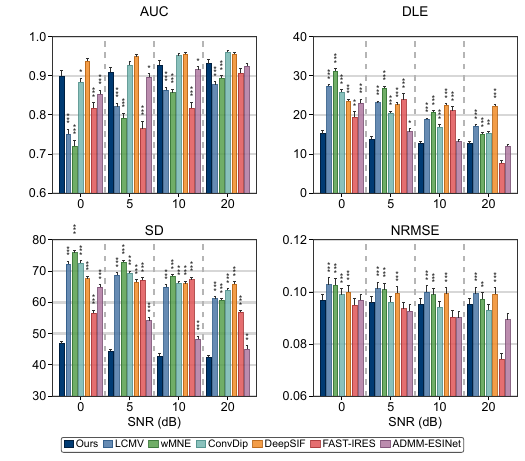}
\caption{Performance metrics for seven ESI methods at SNRs of 0, 5, 10, and 20 dB. Bars show means across 100 Monte Carlo trials per condition, with upper error bars indicating one SEM. Each condition includes mixed single- and two-source configurations. DLE and SD are in millimeters. Asterisks indicate superiority of the proposed method using one-sided paired Wilcoxon tests with Holm correction: *\emph{p}\textless0.05, **\emph{p}\textless0.01, and ***\emph{p}\textless0.001.}
\label{fig:4}
\end{figure}

We evaluated performance under three simulation scenarios with varying SNR levels (0, 5, 10, and 20 dB), source areas (3, 5, and 10 cm$^2$), and source counts (1, 2, 3, and 4), with 100 Monte Carlo trials per parameter setting. The source-area and source-count experiments used an SNR of 10 dB. PINR-ESI was compared with its eLORETA initialization at SNRs of 0, 5, 10, and 15 dB to assess reconstruction gains (\figref{9}). Real-data validation included quantitative assessment against surgical resection regions in the HDEEG-IED-SurgOutcome epilepsy cohort (\figref{10}) and qualitative assessment against known stimulation sites in Localize-MI (\figref{11}).

The simulation experiments used an fsaverage cortical surface, a BioSemi64 EEG montage, and an ico4 source space with 5,124 fixed-orientation cortical vertices. The methods were evaluated on the same forward geometry, source space, and matched simulation cases. \tabref{tab:protocol} summarizes the protocol. Patient-specific forward models for the epilepsy cohort used FreeSurfer cortical surfaces, an ico4 source space, and a three-layer BEM.

\begin{table}[!t]
\caption{Main evaluation protocol}
\label{tab:protocol}
\centering
\footnotesize
\renewcommand{\arraystretch}{1.18}
\renewcommand{\tabularxcolumn}[1]{m{#1}}
\begin{tabularx}{\columnwidth}{@{}>{\raggedright\arraybackslash}m{0.24\columnwidth}X@{}}
\toprule
\textbf{Component} & \textbf{Protocol}\\
\midrule
Source space & fsaverage cortical surface; ico4 cortical source mesh\\
Sensors & BioSemi64 EEG montage\\
Evaluation axes & SNR: 0, 5, 10, 20 dB; source extent: 3, 5, and 10 cm$^2$; source count: 1, 2, 3, 4\\
Test cases & 100 Monte Carlo trials per condition; same cases used for all methods\\
Metrics and tests & AUC and NRMSE (dimensionless); DLE and SD (mm); one-sided paired Wilcoxon signed-rank tests with Holm correction\\
\bottomrule
\end{tabularx}
\end{table}

\subsection{Compared Methods and Implementation Details}\label{b.-compared-methods-and-supervised-fairness}

The comparison included three conventional model-based methods (LCMV, wMNE, and FAST-IRES) and three supervised deep learning methods (ConvDip, DeepSIF, and ADMM-ESINet), together with PINR-ESI. The supervised methods were adapted for single-time-point source imaging. For the simulation experiments, they shared 20,000 simulated EEG--source training pairs generated from 5,000 source configurations at SNRs of 0, 5, 10, and 20 dB. The source configurations covered different source counts and extents. Each supervised method used a single model trained on mixed-SNR data for all simulation settings. For real-data evaluation, the supervised models used training data generated with the corresponding individual forward models. PINR-ESI optimized each observation independently without paired source-label training, with parameter settings fixed within each experimental sweep. It was implemented in PyTorch and run on an NVIDIA GeForce RTX 4090 GPU. In the simulation experiments, Adam was used with a learning rate of $10^{-3}$, with 100 parameter updates for prior-map fitting followed by 300 updates for refinement. The source estimate corresponding to the lowest refinement objective was retained as the final reconstruction.

\begin{figure*}[!tb]
\centering
\includegraphics[width=\textwidth,trim={3.99bp 4.95bp 0.00bp 2.97bp},clip]{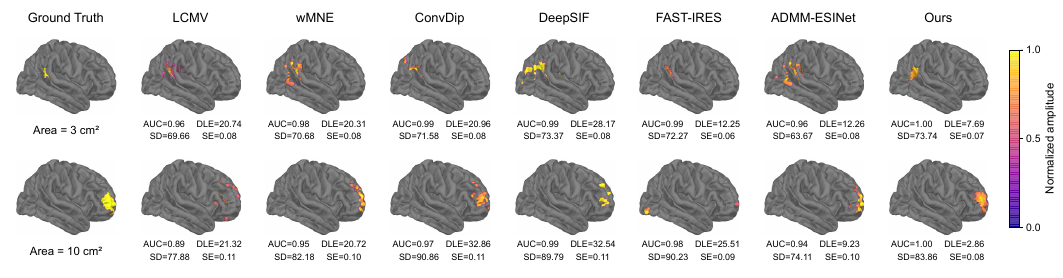}
\caption{ESI results of seven methods with target source areas of 3 cm$^2$ (top) and 10 cm$^2$ (bottom) at 10 dB SNR. Source maps are independently normalized to 0--1. The two single-source examples have different source locations.}
\label{fig:5}
\end{figure*}

\begin{figure}[!tb]
\centering
\includegraphics[width=\columnwidth,trim={11.00bp 0.00bp 0.00bp 0.00bp},clip]{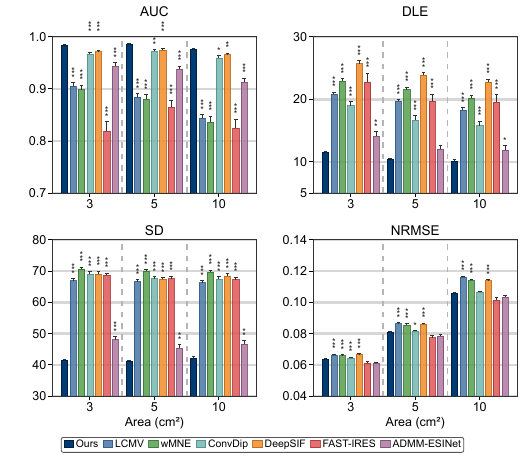}
\caption{Performance metrics for seven ESI methods across target source areas of 3, 5, and 10 cm$^2$. Bars show means across 100 Monte Carlo trials per condition, with upper error bars indicating one SEM. DLE and SD are in millimeters. Asterisks indicate superiority of the proposed method using one-sided paired Wilcoxon tests with Holm correction: *\emph{p}\textless0.05, **\emph{p}\textless0.01, and ***\emph{p}\textless0.001.}
\label{fig:6}
\end{figure}

\subsection{Metrics and Statistical Testing}\label{c.-metrics-and-statistical-testing}

Four metrics were used to evaluate simulated reconstructions: the area under the receiver operating characteristic curve (AUC) for active-region detection, distance localization error (DLE) for localization, spatial dispersion (SD) for spatial spread \cite{ref9}, and energy-aligned full-map normalized root-mean-square error (NRMSE) for reconstruction error \cite{ref17}. A higher AUC and lower DLE, SD, and NRMSE indicate better performance. DLE and SD are reported in millimeters, whereas AUC and NRMSE are dimensionless. DLE was calculated as the mean of the two directional nearest-neighbor distances between the estimated and ground-truth active regions, and NRMSE measured the full-map difference between the energy-aligned estimate and ground truth. For DLE, active regions were identified by applying Otsu's method to min--max normalized source amplitudes. Simulation results were summarized as means and standard errors of the mean (SEM). One-sided paired Wilcoxon signed-rank tests compared PINR-ESI with each baseline, and Holm correction was applied across the six comparisons for each condition and metric. For the epilepsy data, bidirectional DLE between the estimated active region and the surface-projected surgical resection region was used to assess spatial agreement.

\begin{figure*}[!tb]
\centering
\includegraphics[width=\textwidth,trim={4.99bp 5.98bp 0.00bp 2.99bp},clip]{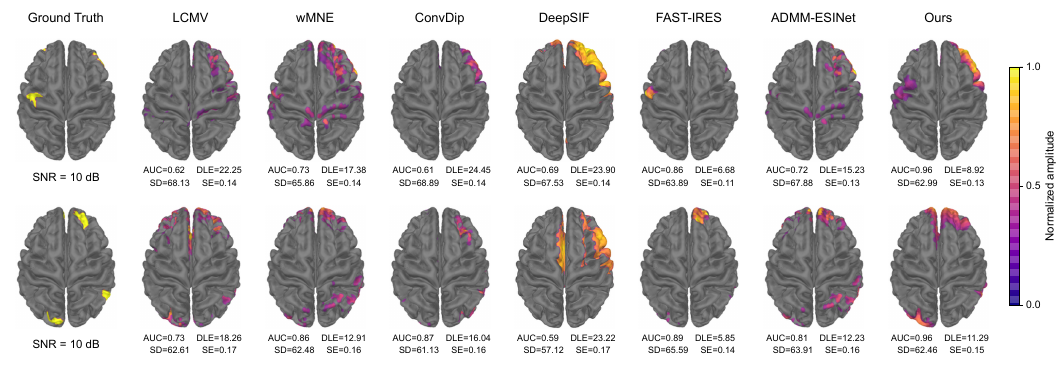}
\caption{ESI results of seven methods with three (top) and four (bottom) source patches at 10 dB SNR. Each source map is independently normalized to 0--1. Values below each reconstruction indicate AUC, DLE (mm), SD (mm), and SE (energy-aligned full-map NRMSE).}
\label{fig:7}
\end{figure*}

\begin{figure}[!tb]
\centering
\includegraphics[width=0.95\columnwidth,trim={11.00bp 0.00bp 0.00bp 0.00bp},clip]{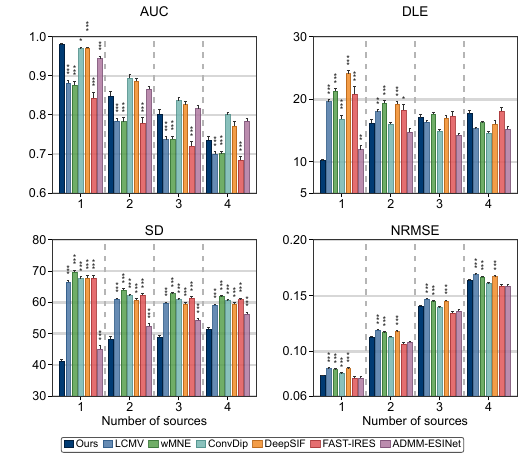}
\caption{Performance metrics for seven ESI methods with one to four sources. Bars show means across 100 Monte Carlo trials per condition, with upper error bars indicating one SEM. DLE and SD are in millimeters. Asterisks indicate superiority of the proposed method using one-sided paired Wilcoxon tests with Holm correction: *\emph{p}\textless0.05, **\emph{p}\textless0.01, and ***\emph{p}\textless0.001.}
\label{fig:8}
\end{figure}

\section{Results}\label{v.-results}

We first evaluate the influence of SNR, source extent, and number of sources through controlled simulations, presenting illustrative reconstructions alongside quantitative results. A paired comparison with eLORETA evaluates the reconstruction gains beyond the initial estimate. We then assess the geometric agreement between estimated active cortical regions and surgical resection regions in 24 epilepsy patients with favorable surgical outcomes.

\subsection{Results of simulation experiments}\label{a.-results-of-simulation-experiments}

\subsubsection{Influence of SNR} \figref{3} shows single-source reconstructions at 5 and 20 dB, and \figref{4} summarizes the performance at 0, 5, 10, and 20 dB. As SNR increased, most methods showed higher AUC and lower DLE and SD, indicating that reduced noise improved source reconstruction. The proposed method achieved the lowest SD at all tested SNRs and the lowest DLE at 0, 5, and 10 dB, indicating good localization and limited spatial dispersion even under relatively noisy conditions. Other methods showed advantages on different metrics: DeepSIF achieved higher AUC, and ConvDip also showed an AUC advantage at some SNRs; FAST-IRES performed better in NRMSE and also achieved a lower DLE at 20 dB. DeepSIF and ConvDip were both trained on simulated data covering multiple SNR levels. This supervised training may help them learn source patterns under the corresponding noise conditions and distinguish active from inactive regions, contributing to their favorable AUC performance.

\subsubsection{Influence of source extent} \figref{5} shows single-source reconstructions with source areas of 3 and 10 cm$^2$, and \figref{6} summarizes the results for 3, 5, and 10 cm$^2$. As source area increased, the AUC of LCMV and wMNE decreased, whereas that of the proposed method remained relatively stable. The proposed method achieved the highest AUC and the lowest DLE and SD at all three source areas, indicating good detection and localization of active regions across source extents. FAST-IRES and ADMM-ESINet had lower NRMSE, showing an advantage in full-map reconstruction. The proposed method's performance may benefit from the combination of a continuous source field and spatial support constraints. The network fits the spatial distribution of source activity, while the prior-guided constraints softly penalize activity outside the prior region, helping to limit spatial dispersion.

\subsubsection{Influence of the number of sources} \figref{7} shows reconstructions with three and four sources, and \figref{8} summarizes the results as the number of sources increased from one to four. Across methods, AUC generally decreased and NRMSE increased with the number of sources. The proposed method showed good detection and localization of active regions in the single-source condition and maintained the lowest SD across all source counts, although its advantages in AUC and DLE diminished with multiple sources. ConvDip, DeepSIF, and ADMM-ESINet achieved higher AUC in the multiple-source conditions, and ADMM-ESINet also showed lower DLE. The consistently low SD of the proposed method may be related to its prior-guided spatial support constraints, which help limit the spread of reconstructed activity. However, the superposition of signals from multiple sources at the sensors still makes it harder to distinguish individual active regions.

\subsubsection{Comparison with the eLORETA initialization}

\begin{figure}[!tb]
\centering
\includegraphics[width=\columnwidth,trim={5.00bp 1.98bp 0.00bp 0.00bp},clip]{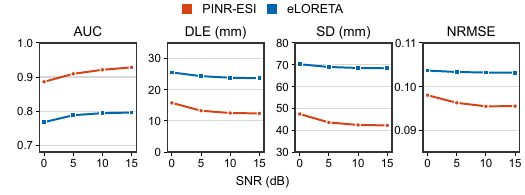}
\caption{Comparison of the complete PINR-ESI reconstruction and the eLORETA estimate used for initialization at SNRs of 0, 5, 10, and 15 dB. Curves show means across 100 source configurations per SNR, with matched cases for both methods. The four panels show AUC, DLE, SD, and NRMSE. DLE and SD are in millimeters.}
\label{fig:9}
\end{figure}

To assess the reconstruction gains of PINR-ESI over a classical inverse solution, we compared the complete method with the eLORETA estimate used for initialization. As shown in \figref{9}, PINR-ESI achieved a higher mean AUC and lower mean DLE, SD, and NRMSE at all tested SNRs, indicating improvements in active-region detection, localization, spatial dispersion, and full-map reconstruction error. eLORETA provides the initial source distribution and spatial support information. PINR-ESI then continues to optimize the source field using EEG consistency and spatial constraints, allowing the reconstruction to be refined beyond the initial estimate. These results support the additional reconstruction benefits of the complete optimization process beyond the classical initialization.

\begin{figure*}[!tb]
\centering
\includegraphics[width=\textwidth,trim={0.00bp 10.99bp 1.99bp 2.99bp},clip]{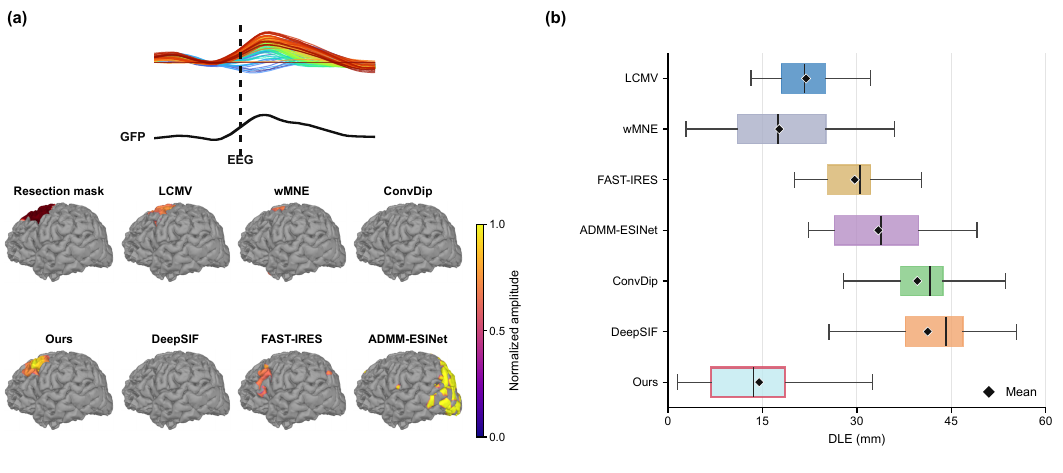}
\caption{Real-data evaluation on the HDEEG-IED-SurgOutcome cohort. (a) EEG and source maps for sub-pat02 at mid-GFP, shown with the RAMPS resection mask. (b) Bidirectional DLE between estimated active cortical regions and the full surface-projected resection region across 24 ILAE-1 subjects. Boxes show medians and interquartile ranges; whiskers follow the 1.5-IQR rule, and diamonds denote means. Outlier symbols are omitted, but all values are included in the statistics.}
\label{fig:10}
\end{figure*}

\begin{figure*}[!tb]
\centering
\includegraphics[width=\textwidth,trim={0.00bp 0.00bp 0.00bp 0.00bp},clip]{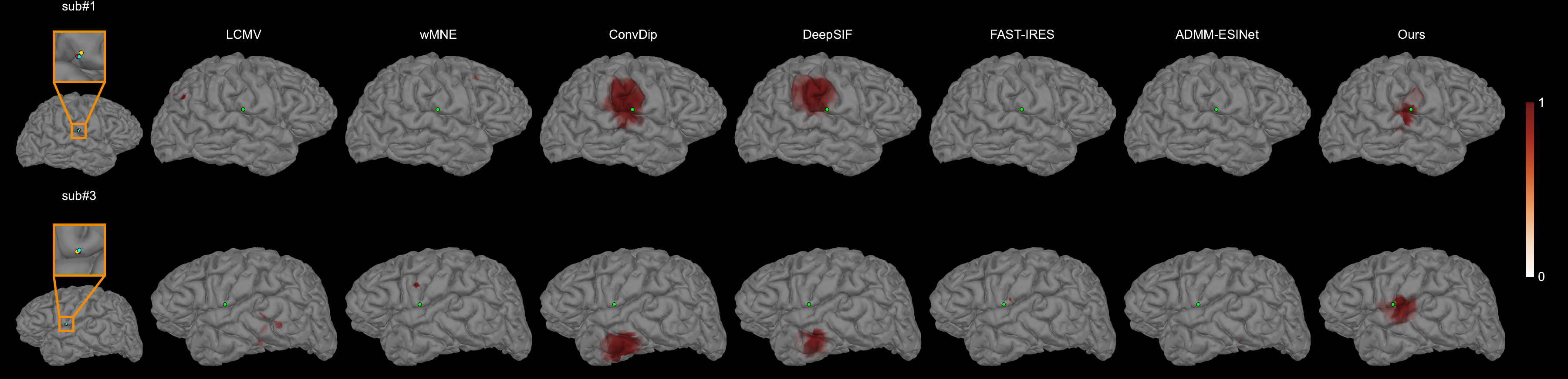}
\caption{Qualitative source imaging results for Localize-MI subjects 1 and 3 in the upper and lower rows, respectively. The insets show the two stimulating contacts as red and yellow points, with their midpoint marked in blue. The projected midpoint is also marked in green on each reconstructed source map. Maps are independently normalized to 0--1 and projected onto the cortical view.}
\label{fig:11}
\end{figure*}

\subsection{Results of Real Data}\label{sec:real}

\subsubsection{HDEEG-IED-SurgOutcome data} We analyzed averaged preoperative interictal epileptiform discharges from 24 patients in HDEEG-IED-SurgOutcome \cite{ref18} who had an ILAE 1 outcome at 12 months after surgery and passed quality control. All methods used patient-specific forward models and reconstructed sources when the GFP of the averaged spike first reached 50\% of its peak. Resection masks generated from pre- and postoperative MRI using RAMPS \cite{ref19} were projected onto the cortical source space and used as clinical spatial references \cite{ref17}. Agreement was assessed by bidirectional DLE between the estimated active regions and resection masks. Supervised baselines were trained on synthetic data generated with patient-specific forward models, while PINR-ESI used wMNE initialization.

\figref[(a)]{10} shows the EEG, resection mask, and reconstructions from seven methods for patient sub-pat02. The main activity reconstructed by PINR-ESI was close to the resection region, whereas some comparators showed more dispersed activity or activity away from this region. Across the 24 patients in \figref[(b)]{10}, PINR-ESI achieved the lowest mean and median DLE. PINR-ESI had a mean DLE of 14.49 mm, followed by wMNE at 17.67 mm. The representative case and cohort results support good spatial agreement between PINR-ESI estimates and surgical resection regions in this cohort.

\subsubsection{Localize-MI data} We further used Localize-MI \cite{ref23} for qualitative evaluation, with the intracranial stimulating contacts and their midpoint as spatial references. Preprocessed trials were averaged within each session, and sources were reconstructed at the time point with maximum GFP within $\pm2$~ms of stimulation. All methods used the same retained channels, individual forward model, and source space; PINR-ESI used eLORETA initialization. \figref{11} shows two selected examples from subjects 1 and 3. In subject 1, ConvDip and DeepSIF produced broader activity near the reference, while PINR-ESI produced a more concentrated reconstruction in its vicinity. In subject 3, the dominant activity from ConvDip and DeepSIF appeared below the reference, whereas that from PINR-ESI was near it. These examples provide qualitative support for the feasibility of PINR-ESI in reconstructing activity near known stimulation sites in real stimulation recordings.

\section{Discussion}\label{vi.-discussion}

PINR-ESI reconstructs extended cortical activity by fitting a continuous neural field to measured EEG without paired source labels or supervised pretraining. Its most consistent benefit in the evaluated simulations was lower spatial dispersion across noise levels, source areas, and source counts (\figsref{4}, \ref{fig:6}, and \ref{fig:8}). At 0 dB, the method achieved the lowest mean DLE and SD, supporting good localization even at the lowest tested SNR. It also achieved the highest AUC and the lowest DLE and SD across the tested source areas. Its relative advantages in AUC and DLE diminished with multiple sources, and several baselines achieved lower NRMSE. These findings indicate that PINR-ESI achieves accurate source localization with low spatial dispersion across the evaluated source configurations.

PINR-ESI combines a continuous INR source-field representation with forward-consistency-guided self-supervised optimization. EEG consistency provides an observation-specific fitting objective, while the neural field shares parameters across spatial locations. FEP samples and locally smooths the field before lead-field mapping. The classical inverse estimate initializes the network and defines support regions for PGSR, providing soft spatial guidance during refinement. In the paired eLORETA comparison (\figref{9}), the complete method improved all four mean metrics under the reported configuration, supporting an additional benefit from the complete reconstruction stage.

The real-data analyses provide complementary evidence. In the 24-patient epilepsy cohort, PINR-ESI achieved lower mean and median resection-based DLE than the compared methods (\figref{10}). In two selected Localize-MI examples, reconstructed activity was near the known stimulation sites (\figref{11}). Together, these results support the feasibility of PINR-ESI for source imaging in clinical and stimulation EEG recordings. Its ability to reconstruct sources without paired source-label training is useful when such data are unavailable or do not capture the characteristics of clinical recordings.

Several limitations remain. The current formulation reconstructs selected EEG time points independently and does not model temporal continuity, which may limit localization accuracy in complex settings involving multiple simultaneous sources. Extending the neural field to the temporal domain and introducing constraints across successive EEG topographies may improve source separation and temporal consistency. Per-observation iterative optimization is also computationally demanding and does not yet support real-time source imaging. More efficient optimization, compact networks, and reuse of initialization across adjacent time points could reduce reconstruction time. In addition, the simulations used simplified source configurations and additive Gaussian noise, which capture only part of the complexity of real EEG recordings. Future studies should include more realistic source dynamics and noise models. Further evaluation should examine negative SNRs, physiological artifacts, inaccuracies in the forward model, and errors in the initial spatial estimate. Together, these directions may improve the temporal consistency and computational efficiency of PINR-ESI and clarify its robustness under more realistic conditions.

\section{Conclusion}\label{vii.-conclusion}

This work presents PINR-ESI, a physics-informed self-supervised framework that combines a continuous INR source-field representation, forward EEG prediction, and prior-guided support regularization to reconstruct extended cortical activity without paired source labels or supervised pretraining. Simulations showed consistent benefits in spatial dispersion control and favorable localization across the evaluated noise levels and source extents, including at 0 dB, with additional reconstruction gains over the eLORETA initialization in the paired comparison. Quantitative evaluation in the epilepsy cohort and qualitative Localize-MI examples supported spatial agreement with surgical resection regions and known stimulation sites, respectively. Future work will focus on temporal modeling, separation of multiple sources, and computational efficiency.


\begin{thebibliography}{23}

\bibitem{ref1}
Z. He, D. Ding, and B. Yuan, ``Electrophysiological source imaging: A noninvasive window to brain dynamics,'' \emph{IEEE Rev.~Biomed. Eng.}, 2018.

\bibitem{ref2}
S. Asadzadeh \emph{et al}., ``A systematic review of EEG source localization techniques and their applications on diagnosis of brain abnormalities,'' \emph{J. Neurosci. Methods}, 2020.

\bibitem{ref3}
A. Sohrabpour and B. He, ``Exploring the extent of source imaging: Recent advances in noninvasive electromagnetic brain imaging,'' Curr. Opin. Biomed. Eng., vol. 18, Art. no. 100277, 2021.

\bibitem{ref4}
H. Sun \emph{et al}., ``Deep neural networks constrained by neural mass models improve electrophysiological source imaging of spatiotemporal brain dynamics,'' \emph{Proc. Natl. Acad. Sci. U.S.A.}, 2022.

\bibitem{ref5}
Z. Feng et al., ``Explaining E/MEG source imaging and beyond: An updated review,'' IEEE J. Biomed. Health Inform., 2025, doi: \href{https://doi.org/10.1109/JBHI.2025.3588350}{10.1109/JBHI.2025.3588350}.

\bibitem{ref6}
A. Sohrabpour \emph{et al}., ``Noninvasive electromagnetic source imaging of spatiotemporally distributed epileptogenic brain sources,'' \emph{Nat. Commun.}, 2020.

\bibitem{ref7}
Z. Feng, C. Guan, and Y. Sun, ``Block-Champagne: A novel Bayesian framework for imaging extended E/MEG source,'' IEEE Trans. Med. Imaging, 2025, doi: \href{https://doi.org/10.1109/TMI.2025.3642620}{10.1109/TMI.2025.3642620}.

\bibitem{ref8}
L. Hecker et al., ``ConvDip: A convolutional neural network for better EEG source imaging,'' Front. Neurosci., vol. 15, Art. no. 569918, 2021.

\bibitem{ref9}
X. Liu \emph{et al}., ``ADMM-ESINet: A deep unrolling network for EEG extended source imaging,'' \emph{IEEE J. Biomed. Health Inform.}, 2025.

\bibitem{ref10}
S. Reynaud \emph{et al}., ``Comprehensive analysis of supervised learning methods for electrical source imaging,'' 2024.

\bibitem{ref11}
M. Morik, A. Hashemi, K.-R. Müller, S. Haufe, and S. Nakajima, ``Enhancing brain source reconstruction by initializing 3D neural networks with physical inverse solutions,'' IEEE Trans. Med. Imaging, 2025, doi: \href{https://doi.org/10.1109/TMI.2025.3594724}{10.1109/TMI.2025.3594724}.

\bibitem{ref12}
A. Molaei \emph{et al}., ``Implicit neural representation in medical imaging: A comparative survey,'' 2023.

\bibitem{ref13}
J. Barron \emph{et al}., ``Mip-NeRF: A multiscale representation for anti-aliasing neural radiance fields,'' \emph{ICCV}, 2021.

\bibitem{ref14}
E. Najaf and G. Ongie, ``Towards a sampling theory for implicit neural representations,'' 2024.

\bibitem{ref15}
Y. Chen \emph{et al}., ``Anti-aliasing projection representation field for inverse problems in imaging,'' 2023.

\bibitem{ref16}
S. Essakine \emph{et al}., ``Where do we stand with implicit neural representations? A technical and performance survey,'' 2024.

\bibitem{ref17}
S. Wang \emph{et al.}, ``A geometry-aware framework enhances noninvasive mapping of whole human brain dynamics,'' Nat. Biomed. Eng., 2026, doi: \href{https://doi.org/10.1038/s41551-026-01664-0}{10.1038/s41551-026-01664-0}.

\bibitem{ref18}
B. J. Vorderwülbecke \emph{et al.}, ``High-density EEG source localisation of averaged interictal epileptic discharges validated by surgical outcome,'' Sci. Data, vol. 12, Art. no. 1441, 2025.

\bibitem{ref19}
C. Simpson, G. Hall, J. S. Duncan, Y. Wang, and P. N. Taylor, ``Automated generation of epilepsy surgery resection masks: The RAMPS pipeline,'' Imaging Neurosci., vol. 3, Art. no. IMAG.a.147, 2025.

\bibitem{ref20}
M. Jiao et al., ``XDL-ESI: Electrophysiological sources imaging via explainable deep learning framework with validation on simultaneous EEG and iEEG,'' NeuroImage, vol. 299, Art. no. 120802, 2024, doi: \href{https://doi.org/10.1016/j.neuroimage.2024.120802}{10.1016/j.neuroimage.2024.120802}.

\bibitem{ref21}
L. Wang et al., ``VSSI$_{2p}$-Net: Physics-guided deep unfolding with $\ell_{2,p}$-norm and variation sparsity for EEG source imaging,'' NeuroImage, vol. 328, Art. no. 121793, 2026, doi: \href{https://doi.org/10.1016/j.neuroimage.2026.121793}{10.1016/j.neuroimage.2026.121793}.

\bibitem{ref22}
G. Huang et al., ``Electromagnetic source imaging via a data-synthesis-based convolutional encoder--decoder network,'' IEEE Trans. Neural Netw. Learn. Syst., vol. 35, no. 5, pp. 6423--6437, 2024, doi: \href{https://doi.org/10.1109/TNNLS.2022.3209925}{10.1109/TNNLS.2022.3209925}.

\bibitem{ref23}
E. Mikulan et al., ``Simultaneous human intracerebral stimulation and HD-EEG, ground-truth for source localization methods,'' Sci. Data, vol. 7, Art. no. 127, 2020, doi: \href{https://doi.org/10.1038/s41597-020-0467-x}{10.1038/s41597-020-0467-x}.

\end{thebibliography}
\end{document}